\documentclass[12pt, a4paper]{elsarticle}
\usepackage[margin=1in]{geometry}
\usepackage{amssymb,amsmath}
\usepackage[english]{babel}
\usepackage{inputenc}
\usepackage{makecell}
\usepackage{multirow,multicol}
\usepackage{accents}
\usepackage{algorithm}
\usepackage{algorithmicx}
\usepackage{algcompatible}
\usepackage{algpseudocode}
\usepackage{setspace}

\usepackage[figurename=Fig.,font=small,skip=1 pt]{caption}
\graphicspath{ {./Image_6/}}
\newcommand{\justified}{%
  \rightskip\z@skip%
  \leftskip\z@skip}
\usepackage{enumerate}
\usepackage{url}
\usepackage{color, soul, colortbl}

\journal{Mechanical Systems and Signal Processing}

\begin{document}

\begin{frontmatter}




\title{AI-driven Inverse Design of Band-Tunable Mechanical Metastructures for Tailored Vibration Mitigation}


\author{Tanuj Gupta$^a$, Arun Kumar Sharma$^a$, Ankur Dwivedi$^a$, Vivek Gupta$^b$, Subhadeep Sahana$^a$, Suryansh Pathak$^c$, Ashish Awasthi$^a$, and Bishakh Bhattacharya$^{a}$} 

\affiliation{organization={Department of Mechanical Engineering, Indian Institute of Technology Kanpur},
            postcode={208016}, 
            country={India}}

\affiliation{organization={Department of Precision and Microsystems Engineering, Delft University of Technology},
            postcode={2628 CD}, 
            country={Netherlands}}

\affiliation{organization={Department of Mechanical Engineering, Motilal Nehru National Institute of Technology Allahabad},
            city={Prayagraj},
            postcode={211004}, 
            country={India}}

\begin{abstract}
On-demand vibration mitigation in a mechanical system needs the suitable design of multiscale metastructures, involving complex unit cells. In this study, immersing in the world of patterns and examining the structural details of some interesting motifs are extracted from the mechanical metastructure perspective.  Nine interlaced metastructures are fabricated using additive manufacturing, and corresponding vibration characteristics are studied experimentally and numerically. Further, the band-gap modulation with metallic inserts in the honeycomb interlaced metastructures is also studied. AI-driven inverse design of such complex metastructures with a desired vibration mitigation profile can pave the way for addressing engineering challenges in high-precision manufacturing. The current inverse design methodologies are limited to designing simple periodic structures based on limited variants of unit cells. Therefore, a novel forward analysis model with multi-head FEM-inspired spatial attention (FSA) is proposed to learn the complex geometry of the metastructures and predict corresponding transmissibility. Subsequently, a multiscale Gaussian self-attention (MGSA) based inverse design model with Gaussian function for 1D spectrum position encoding is developed to produce a suitable metastructure for the desired vibration transmittance. The proposed AI framework demonstrated outstanding performance corresponding to the expected locally resonant bandgaps in a targeted frequency range.
\end{abstract}





\end{frontmatter}



\section{Introduction}
\label{sec1}

\begin{figure}[ht!]
		\centering
		\includegraphics[width=\textwidth]{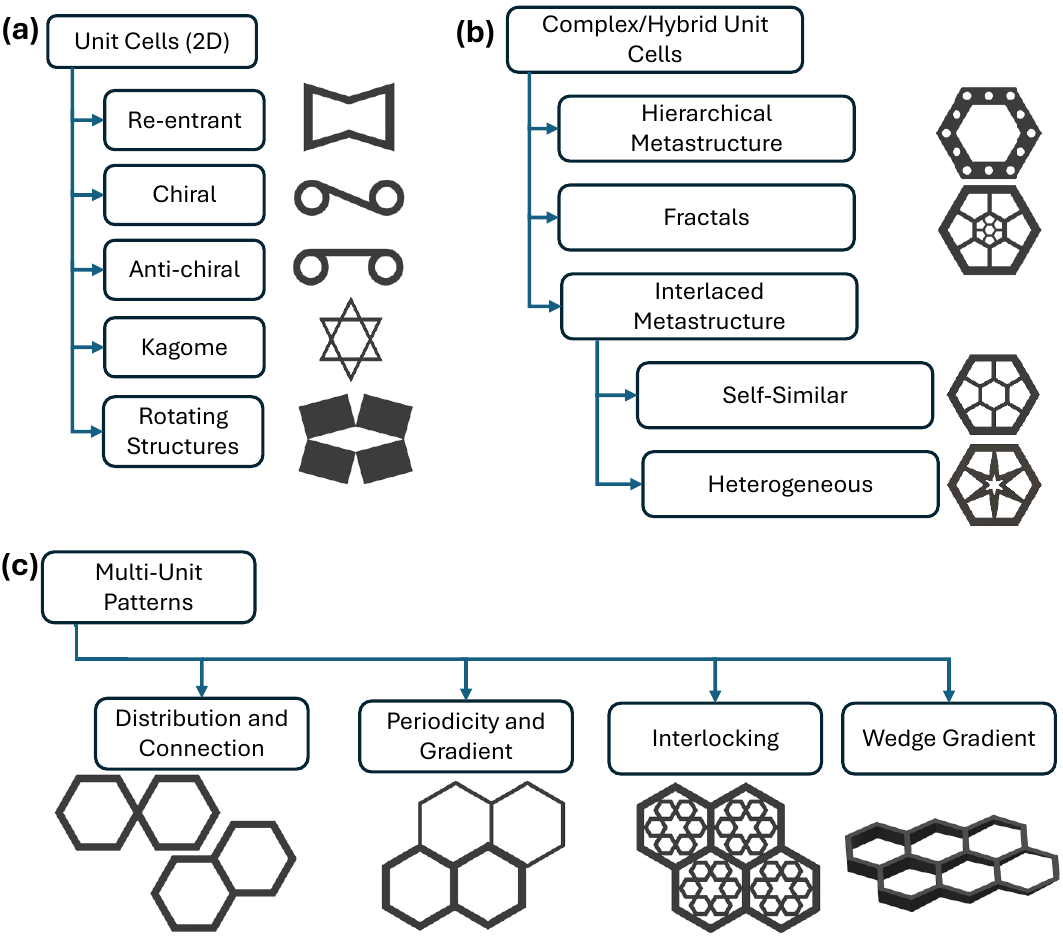}
		\caption{Different types of metastructure unit cells can be categorized based on their lattice structures. (a) Five popular unit cells are used as primary lattices. (b) A complex array of primary lattices can be used to create different possible interesting metastructure unit cells. (c) Multi-unit patterns represent the arrangement of unit cells to form complete 2D and 3D metastructures.}
		\label{fig:Tree_diagram}
\end{figure}

Metamaterials are emerging from varied research domains like electromagnetic, mechanical, and acoustic applications having unique tailored properties that are primarily dependent on their structural geometry rather than intrinsic material properties. It wasn't until the late 19th century, when fundamental physics became better understood, that the development of systematic engineering principles made it possible to explore counter-intuitive physical characteristics of some patterns in periodic structures, generally known as metamaterials and metastructures. Today, it is often proposed as a panacea for controlling attenuation and wavefront divergence related to elastic and electromagnetic wave propagation and there exists a vast literature based on meticulously designed lattice structures with special geometric patterns \cite{wan2024novel,Guancong,silverberg2014using, dwivedi2023line,ai2017metamaterials,wu2025metamaterial}. Generally, these structures have sub-wavelength features designed in the form of lattices on a smaller scale than the wavelength of external physical stimuli they interact with. These engineered lattices may introduce exotic mechanical and electromagnetic properties, manipulating the behavior of electromagnetic waves \cite{horsley2023theory}, elastic waves as well as sound waves \cite{panahi2021novel,sugino2016mechanism,liu2011wave}. As metastructures have structural-dependent properties, hence by carefully designing these lattices, scientists can achieve customizable responses with extraordinary effects like negative stiffness \cite{dwivedi2020simultaneous,YU2023104799,Xiaoming}, negative mass \cite{dwivedi2021optimal}, negative Poisson's ratio \cite{AI201770,singh2022active}, negative thermal expansion coefficient \cite{ZHANG2024108692}, negative reflection \cite{Yihang,Pendry} for wave alteration \cite{Goffaux}, sub-wavelength bandgap \cite{zhao20223d, D'Alessandro,dwivedi2024bandgap} and distinctive electro-dynamic phenomena \cite{Sarychev,chen2024bandgap, Jiao}. The study of lattice geometries in metamaterials opens a pathway to myriad innovative applications ranging from optics to acoustics such as wearable devices \cite{li2020metamaterials,yigci2024ai}, advanced space structures \cite{Brendel,chuang2021multi}, next-gen automobiles \cite{JiaZian,chen2022closed}, high precision robotic manipulators \cite{sun2023embedded}, bionic grippers, \cite{tawk20223d}, anti-seismic structures \cite{lim2019built,guner2024seismic}, and vibration sensing \cite{zhang2024metamaterial}.
Mechanical properties of metastructures designed with simple unit cells such as honeycombs with positive and negative re-entrant angles, chiral, and kagome have been extensively studied through experimental avenues, numerical simulations, and mathematical analyses \cite{HourglassSR,akash2024data,YANG2015475,jiang,XU2022111866,dwivedi2020dynamics,TAKAHIRO}. The generation of a large range of novel periodic structures at different length scales has ushered in phenomenal developments of metamaterials as depicted in Fig. \ref{fig:Tree_diagram}. 

\begin{figure}[H]
		\centering
		\includegraphics[width=\textwidth]{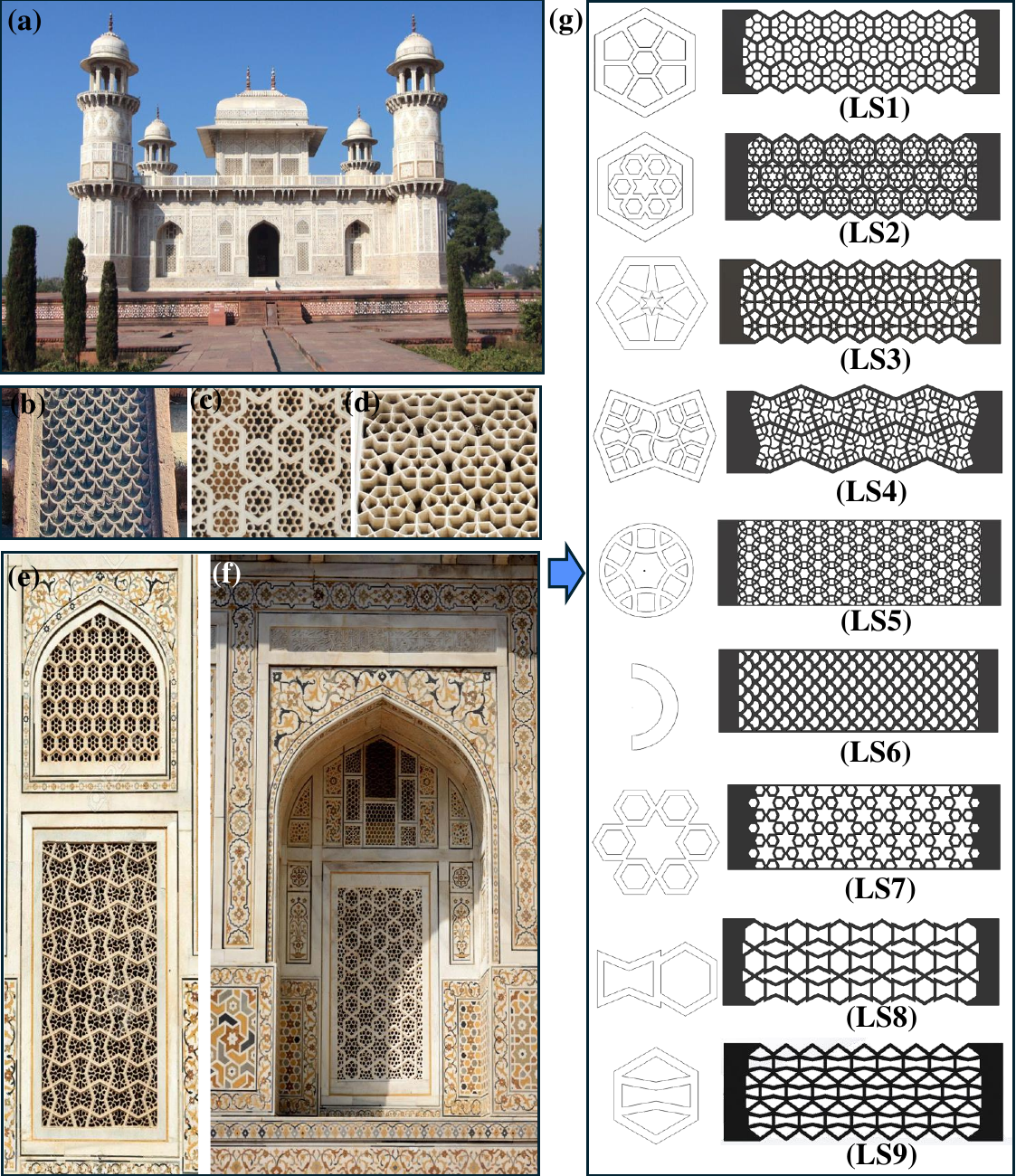}
        \vspace{0.1cm}
		\caption{\textbf{(a)}The Tomb of Itmad-ud-Daula is a mausoleum in the city of Agra in the Indian state of Uttar Pradesh (27.1929 $^{\circ}$ N, 78.0310 $^{\circ}$ E). The architecture of this monument is unique in terms of crafting which uses a combination of pietra-dura and calligraphy-based decorative art techniques. \textbf{(b to d)}. Consequently,  the latticework is depicted through the carved panels of calligraphic design.  The walls, floors, and niches are decorated with beautiful geometrical patterns. \textbf{(e and f)}. Here, we have specifically focused on the Jali screens at the centre of the niches all around the tomb. They were used extensively for elegance and their ability to facilitate the entry of natural light and air inflow. Often it is also used as part of a passive cooling system. \textbf{(g)} Shows the corresponding metastructure models (LS1-LS9) inspired from these architectural elements.}
		\label{fig:tomb}
\end{figure}

Architectural patterns in metamaterials can also play a pivotal role in manipulating electromagnetic and acoustic waves. Intricate architectural patterns have a long history, with certain examples dating back to the 4th century \cite{leonhardt2007invisibility}. These early forms were the outcome of creative heuristic advancements of pattern generation often driven by culture and religion. Geometric design and pattern have played a significant role in the spiritual traditions of many religions.
Considered as the inspiration for the Taj Mahal, the tomb of Itmad-ud-Daula, also known as the `Baby Taj Mahal,' is a stunning example of Indian medieval architecture as shown in Fig. \ref{fig:tomb} (a). This mausoleum displays a wonderful variety of geometric arabesques. 
Such intricate patterns, popularly known as `Jali' in India, were crafted for aesthetic purposes and also as a functional component in passive cooling and ventilation. Some of the patterns are shown in Fig. \ref{fig:tomb}(b-f) with the corresponding metastructural models shown in Fig. \ref{fig:tomb}(g). The related lattice structures can be classified according to different types of unit cells, considering angle, arrangement, and symmetry (Fig. \ref{fig:Tree_diagram}(a)). We investigate various complex/hybrid arrangements of these unit cells (Fig. \ref{fig:Tree_diagram}(b)) and then demonstrate how different arrangements can be used to generate complete metastructures (Fig. \ref{fig:Tree_diagram}(c)). These metastructures provide a foundation for further studies evaluating their performance in controlling elastic wave propagation and vibration.

When we delve into the realm of complex hybrid unit cell-based metastructure it requires advanced manufacturing techniques like 3D printing. Recent advancements in precise additive manufacturing techniques at the microscale have further enabled the fabrication of new lattices with complex topology and interlaced geometry, which are challenging to achieve using traditional subtractive manufacturing methods. Such metastructures have distinct behaviors when subjected to external stimulus such as mechanical excitation \cite{arretche2018interrelationship,dwivedi2022bandgap,aguzzi2022octet, gupta2024tailoring, mirani2023tailoring}. Additive manufacturing has the potential to fabricate precisely complex cellular and architectural metamaterials by manipulating multimaterials with controllable properties. The design and fabrication of smart material-based programmable mechanical metamaterials with 4D printing technology can change their properties dynamically in response to external stimuli. In recent years, there has been a growing interest in developing engineered materials using a range of 3D and 4D additive manufacturing techniques \cite{samal20234d,van20214d}. Such metastructures with complex multiscale geometry are difficult to design according to a desired frequency response with a targeted bandgap. It may require numerous attempts of FEM-based simulations and design changes to achieve the targeted frequency response and, hence, may not be feasible in practice due to high computational costs. Therefore, efficient AI-based models are necessary that can predict the effective design of a metastructure with complex geometry and the desired frequency response.

In recent years, various attempts have been made to develop AI-based inverse design models that can generate complex multiscale mechanical metamaterials having tailored mechanical response and properties \cite{mei2025reconfigurable,zheng2023deep,  chai2024tailoring}. Deng \textit{et. al.} \cite{deng2022inverse} suggested a neural accelerated evolutionary strategy that can obtain a geometry of metamaterials with hinged quadrilaterals for a wide range of target nonlinear mechanical responses.

In another similar work, Ha \textit{et. al.} \cite{ha2023rapid} developed a rapid inverse design method using generative machine learning to generate metamaterials having uniaxial compressive stress-strain response. Recently, Bastek \textit{et. al.} \cite{bastek2023inverse} proposed video-denoising diffusion models that can produce materials with tailored complex target performance. Their AI-based model, having spatial and temporal attention, enables the inverse design of metamaterial with a highly non-linear stress-strain response that closely matches with the finite element simulations. These research works were dedicated to the inverse design of metamaterials with tailored nonlinear stress-strain properties. Similarly, various researchers have also investigated the AI-based inverse design of metamaterials and metasurfaces like non-periodic 3D architectures having predictable anisotropic properties\cite{deng2024ai}, low-index-contrast structures on a silicon photonics platform \cite{nikkhah2024inverse} and antenna surface design \cite{gupta2023tandem}. All these studies have generally focused on the inverse design of unit cells of metamaterials or periodic surfaces for a tailored mechanical or electrical response. 

The inverse design of mechanical metastructures having complex interlaced geometry for tailored frequency response in a defined attenuation bandwidth still remains a challenge due to hierarchical multiscale geometry. Therefore, through this study, we propose a FEM-inspired geometric attention mechanism that can learn the geometry having multiple scales in a metastructure and produce the corresponding frequency response. Further, with the help of the forward model, an inverse design model is developed that can generate a suitable metastructure for the desired vibration response. As mechanical metamaterials have structure-dependent properties hence, their inverse design can revolutionize various application domains with customizable responses by tuning mechanical properties such as stiffness deformation, flexibility, and vibration response. Hence AI-based inverse design of mechanical metamaterials can enable advanced functionalities in prosthetics, exoskeletons, and bone implant designs, dynamic and adaptive structure design in robotics, impact and shock energy absorption lightweight metastructures, and soft robotics \cite{shah2021soft}.

\section{Numerical Simulations}
Finite element analysis-based simulation has been carried out using Ansys, employing the Ansys Parametric Design Language (APDL) solver for all 720 structures, and the host material is chosen as PLA (detailed properties given in Supplementary Table \textcolor{blue}{S3}). The entire structure was discretized by using tetrahedron elements with a 2 mm element size, to ensure convergence for interlaced geometries. The smaller edges of the base structure were fixed, while the longer edges remained free. A dynamic force of amplitude equivalent to 10 N was applied to the base, and the response was calculated on the opposite phase. Modal analysis was performed to identify all modes up to 10 KHz frequency, followed by harmonic analysis to obtain the transmissibility. The mode superposition method was used to link the modal analysis results with the harmonic analysis input. Frequency response functions (FRFs) were generated over the frequency range of 0 to 10 KHz with a 10 Hz frequency resolution; this step size is preferable to accurately capture the response. A structural damping ratio of 0.001 was incorporated to account for the viscoelastic effect of PLA substrate.

\subsection{Dynamic characterization and determination of the Bandgap}

\begin{figure}[ht!]
		\centering
\includegraphics[width=1\textwidth]{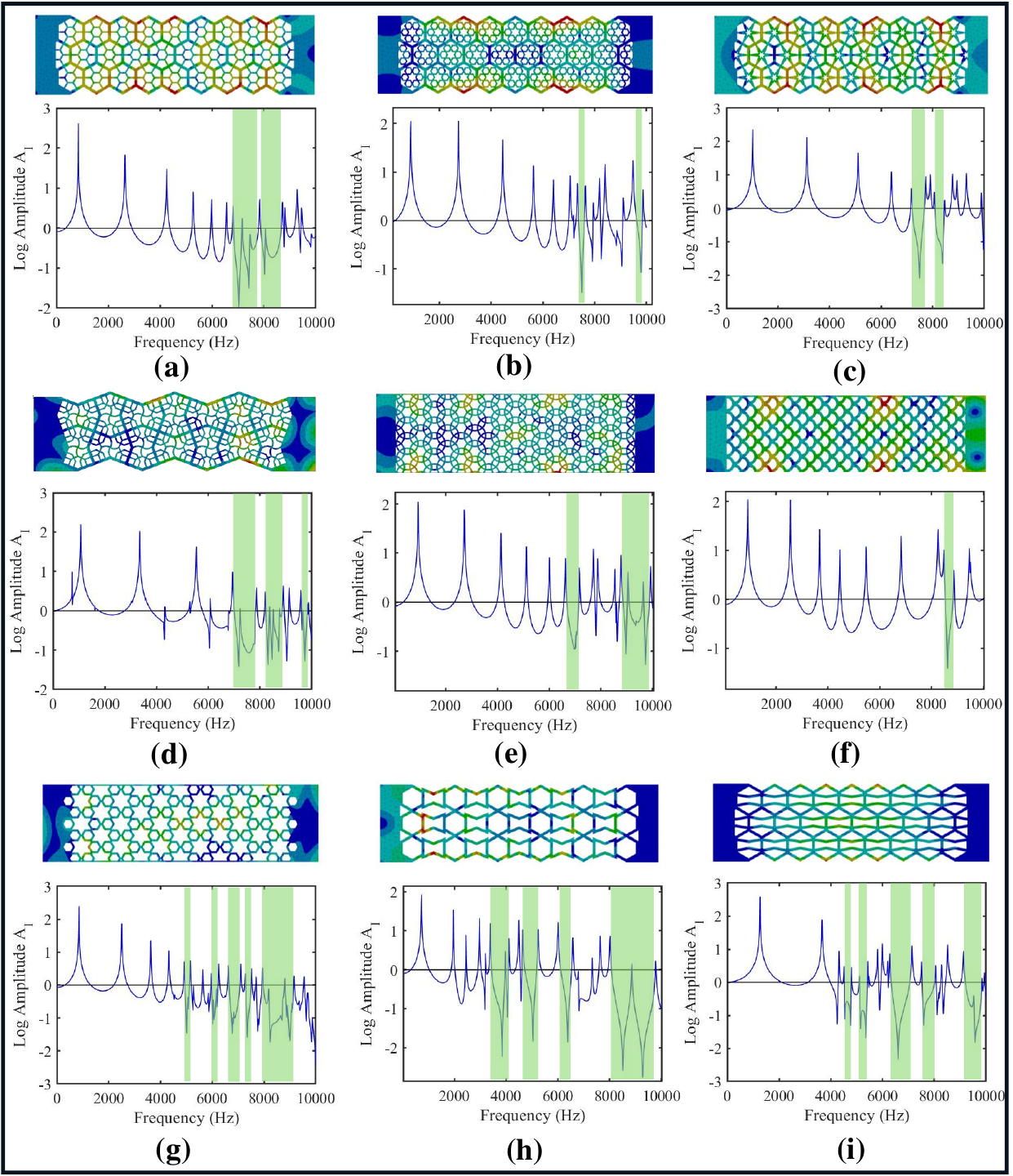}
		\caption{Schematic diagram illustrates latticed mechanical metastructures featuring different interlaced lattice structures (LS1-LS9). Dynamic analysis of the metastructures is performed and mode shapes of lattices at a frequency corresponding to minimum force transmissibility for in-plane excitation are shown. In mode shapes, red indicates maximum deformation and blue indicates minimum deformation.  The attenuation corresponding to each geometry is obtained by using the transmissibility plot for each structure. We have highlighted distinct attenuation zones within the transmissibility response of metastructures through green-shaded patches.}
		\label{fig:all_latices}
\end{figure}

Selected latticed metastructures as shown in Fig. \ref{fig:all_latices}, generated based on architectural elements (described in Fig. \ref{fig:tomb}) have an overall size of 204 mm x 58 mm and a thickness of 5 mm.
\begin{figure}[!ht]
		\centering
		\includegraphics[width=0.84\textwidth]{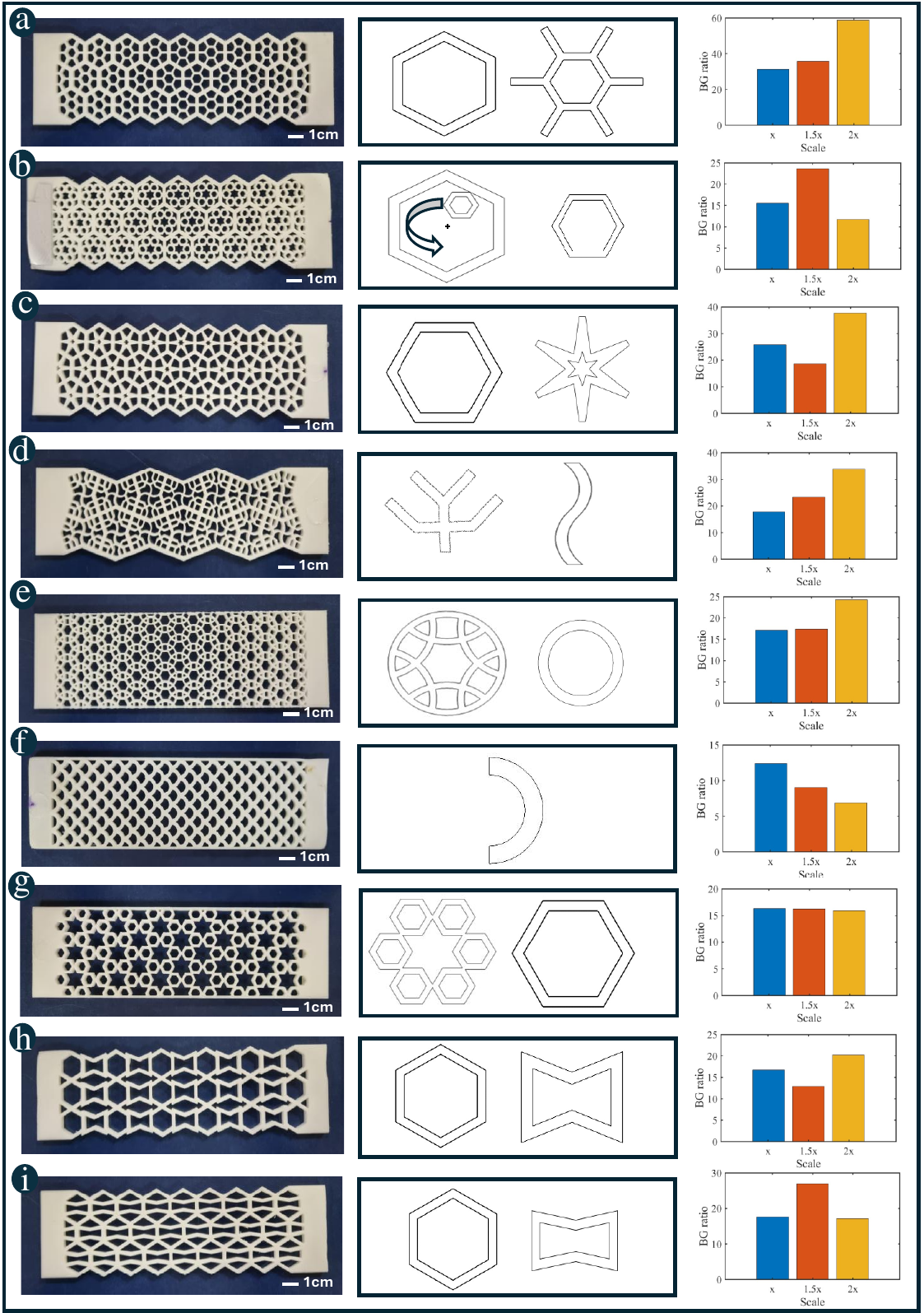}
		\caption{Conducting a comparative analysis, this study investigates the impact of scaling (x, 1.5x, 2x) of the unit cell on the bandgap, with the resulting BG ratio illustrated in a bar diagram. The basic building blocks of unit and sub-unit cells are represented in the middle column of the figure. Subunit cells are used as an interlaced component for the primary structure from which each configuration is constructed. The patterns of latticed mechanical metastructures are 3D Printed using PLA having different primary lattice structures namely (a) Honeycomb-honeycomb interlaced (LS1) (b) Honeycomb interlaced with a honeycomb pattern (LS2) (c) Honeycomb interlaced with a star (LS3) (d) Chiral interlaced structure (LS4) (e) Ring interlaced structure (LS5) (f) Fishtail structure (LS6) (g) Star-honeycomb structure (LS7) (h) Honeycomb-auxetic structure (LS8) (i) Honeycomb-auxetic interlaced structure (LS9). }
		\label{fig:scale_up}
\end{figure}

The size was carefully chosen keeping in mind the manufacturing constraints as well as the requirement of minimum number of periodic units (greater than 7 units) in the tessellated structure to express the metamaterial behavior \cite{dwivedi2021optimal}. On both sides, 15 mm solid frames are appended with the latticed structure in the longitudinal direction (see Supplementary Fig. \textcolor{blue}{S1}) for ease of clamping and excitation during dynamic testing. 
For each latticed structure, there is a frequency band in which the structure may work as an attenuator signifying the absence of wave propagation, and the same is highlighted in Fig. \ref{fig:all_latices}. We have shown the related mode shapes obtained using Ansys corresponding to minimum force transmissibility within the attenuation zone. Different structures have different damping ratios depending on the lattice geometry. The damping ratio provided is calculated by the half-power bandwidth method (highest for LS1 and the lowest for LS9), which is in the supplementary Fig. \textcolor{blue}{S10}. 

 Latticed metastructures with different sub-lattice structures are shown in Fig. \ref{fig:scale_up}. Figure \ref{fig:scale_up}(a-c) illustrates the primary unit cells as honeycomb sub-lattice, single honeycomb, circular pattern of honeycomb cells and star structure respectively. In Figure \ref{fig:scale_up}(d), the primary unit cell has a chiral structure with sub-lattices exhibiting a spider web-like pattern. Furthermore, Fig. \ref{fig:scale_up}(e) features a primary unit cell as a ring structure interlaced with other ring structures. Figure \ref{fig:scale_up}(f) is a fishtail structure. Fig. \ref{fig:scale_up}(g) shows the primary unit cell as a series of honeycombs arranged to form a star at the center of the structure. Figure \ref{fig:scale_up}(h, i) are combinations of honeycomb and auxetic structure. In the former, these structures are present in a series combination, while in the latter, the auxetic structure is interlaced with the honeycomb structure.
The dimensions of the unit cells are detailed in Supplementary Table \textcolor{blue}{S1}. After creating all the metastructures of specified dimensions, we scaled them up by 1.5 and 2 respectively to capture the scale effect of unit cells in the wave transmission pattern. The scaling effect as shown in Fig. \ref{fig:scale_up} is illustrated with the help of the Band Gap (BG) ratio in the bar diagram.

\subsection{Bandgap characteristics of honeycomb-based interlaced structures}
  \begin{figure}[ht!]
		\centering
		\includegraphics[width=\textwidth]{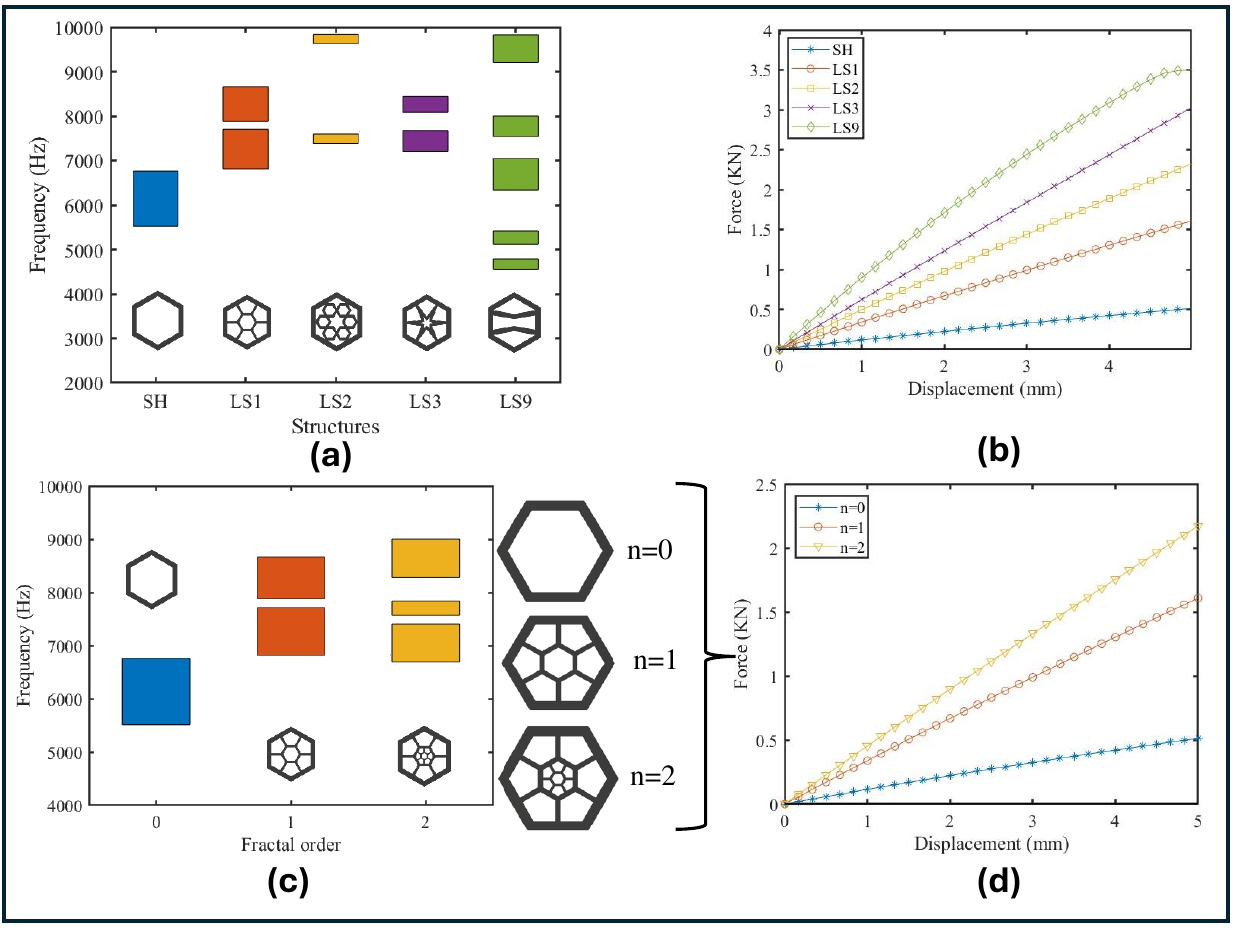}
		\caption{(a) A comparison of bandgap at different frequency levels of different structures interlaced in the honeycomb structure. (b), (d) The comparison of static performance is obtained through analytical computations of different structures. The displacement control mechanism is set as 5 mm compression of the initial height of samples (c) Comparison of bandgap in different order (n=0, 1, 2) of the fractal structure.}
		\label{fig:half_power_BW}
\end{figure}
Comparative analysis of the effect of different interlaced structures on bandgap shift (Fig. \ref{fig:half_power_BW}(a)) is carried out in various frequency ranges. The band gap occurs at a frequency band comparatively higher than that of the simple honeycomb structure. Notably, for the LS3 structure bifurcation in bandgap occurs, and for the last structure (LS9) multiple bandgaps occur both for the higher and lower frequency ranges. Furthermore, we have compared the bandgap range of various interlaced honeycomb unit cells (Fig. \ref{fig:half_power_BW}(b)) along with the force-deflection relationships. The force-deflection relationships are linear for all the patterns except for LS9. 

Figure \ref{fig:half_power_BW}(c) illustrates the fractal structure of different order (n=0, 1, 2) and their corresponding bandgap shifts. In different structures, as the fractal order increases, the bandgap shifts to higher frequency ranges, and for further increase in fractal order multiple bandgaps occur. Furthermore, on the right side we have represented the force-deflection relationship for the fractal structure (see Fig. \ref{fig:half_power_BW}(d)). With the increase in the fractal order, the structure becomes stiffer, with 196.32\% increase from zeroth to first-order fractal and 33.61\% increase from first to second-order fractal. From this study, we can conclude that with increasing the structural complexity inside the structure, the bandgap starts to shift towards the higher frequency range, and for higher-order fractals, it also bifurcates which is verified by taking different fractal structures inside the honeycomb and explained in the Supplementary Fig. \textcolor{blue}{S8}.

\section{Experimental testing}
\begin{figure}[!ht]
		\centering
            \includegraphics[width=16cm]{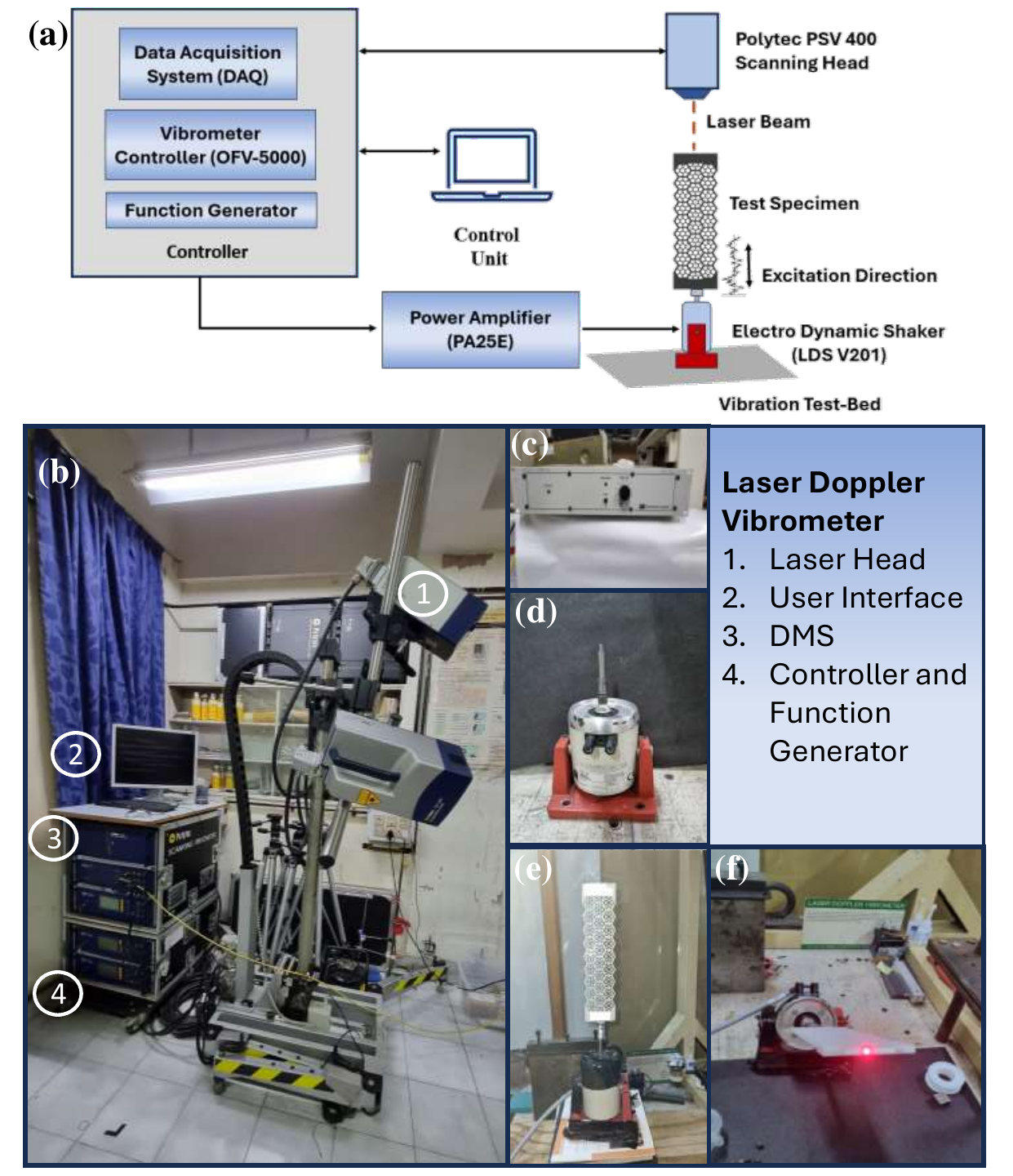}
		\caption{(a) The experimental setup, accompanied by the Laser Doppler Vibrometer (LDV), is depicted in the schematic diagram. (b) 3D laser head with a controller is featured. (c) Power amplifier (d) Electro-dynamic shaker which is employed in setup (e)-(f) Excitation is provided from the base through an electrodynamic shaker, while the response is obtained by Laser Doppler Vibrometer focused on the top surface of the work-piece.}
  \label{fig:experimental_setup}
\end{figure}
\begin{figure}[!ht]
		\centering
		\includegraphics[width=\textwidth]{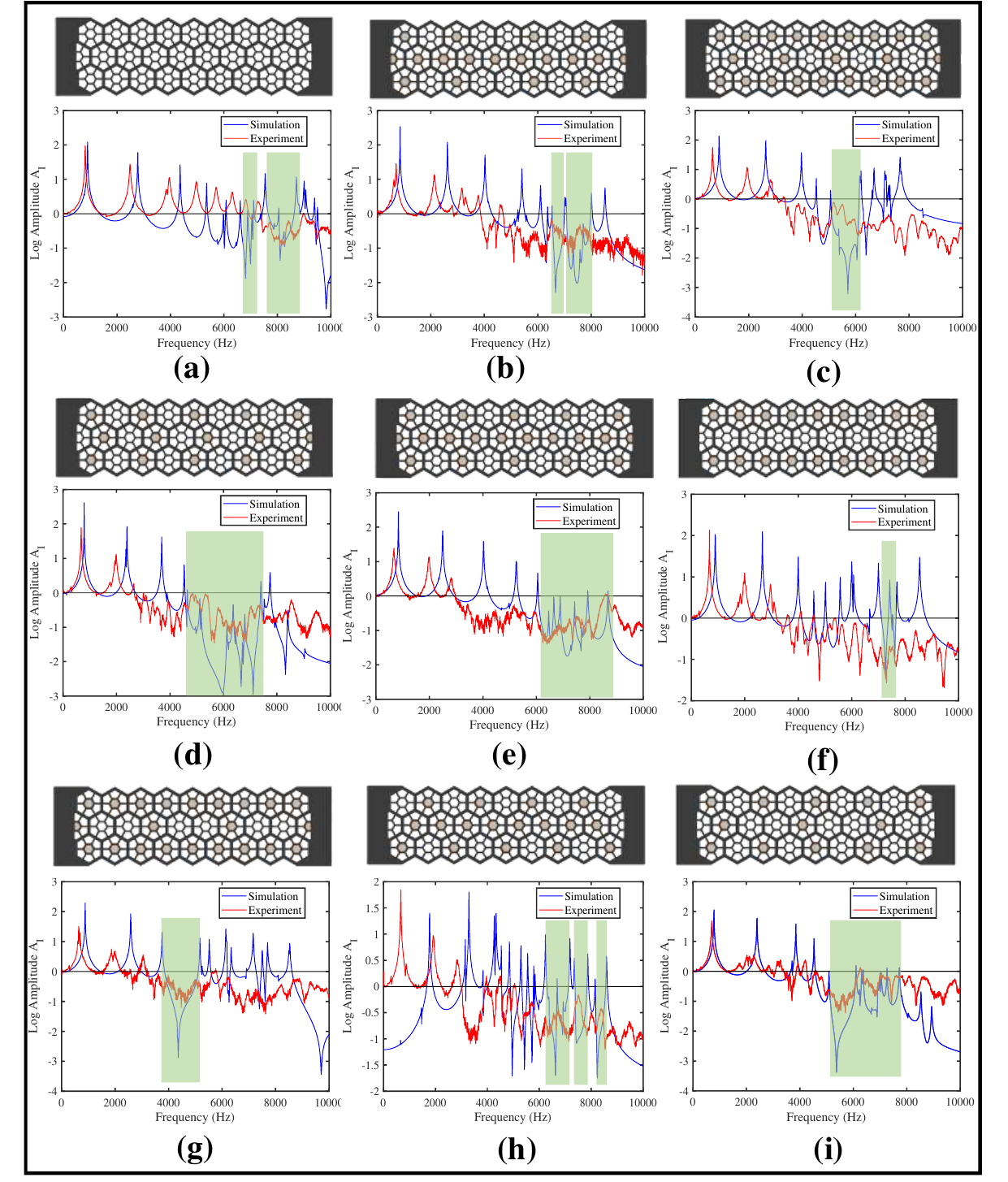}
		\caption{Experimental validation of computational result with variations in mass inserts pattern is shown, and the corresponding bandgaps are marked by green patches. \textbf{(a)} Simple interlaced honeycomb structure (LS1) \textbf{(b)} Diamond pattern with a broken symmetry at the middle \textbf{(c)} Cross pattern \textbf{(d)} Forward arrow pattern \textbf{(e)} Diamond pattern \textbf{(f)} Top and bottom row are filled middle row is empty \textbf{(g)} Top and bottom rows are filled with mass inserts while the middle row has two-gap arrangement with no inserts \textbf{(h)} Upward pointing arrow pattern \textbf{(i)} Top and bottom row inserts are arranged alternatively while middle row has two-gap arrangement with no inserts.}
		\label{fig:experimental_validation}
\end{figure}
The study utilizes a non-contact technique to measure vibration to assess the response of different latticed metastructures. 3D printed samples, each showing unique metastructure, are subjected to dynamic testing using a 3D Laser Doppler Vibrometer (LDV) from Polytec as shown in Fig. \ref{fig:experimental_setup}. 3D printed structures are made of PLA the detailed properties of the material and elemental composition of the hexagonal shape inserts made of mild steel are shown in supplementary Tables \textcolor{blue}{S3} and \textcolor{blue}{S4}. These samples are attached to an Electrodynamic shaker system in cantilever mode. The laser head of LDV captures surface velocity by analyzing the phase shift between the incident and reflected laser beams. Data acquisition and signal processing are executed through the NI-DAQ system. The base excitation technique is utilized to determine transmissibility. In this method, the LDS shaker generates 3200 Fast Fourier Transform (FFT) lines, and the pseudo-random excitation with a frequency range of 0 to 10 KHz. Displacements of the top and bottom surfaces of the samples are recorded to calculate transmissibility after post-processing of the data. A retro-reflective tape is also attached to the surface where data has been recorded to reduce the scattering of the laser beam from the surface and three closely spaced points were taken on both the top and bottom surface for averaging of data to reduce error.

\subsection{Performance measurement of lattice with resonators}
In figure \ref{fig:experimental_validation}, we have selected a primary honeycomb lattice structure interlaced with a single honeycomb unit (LS1) at the centre. Our objective is to study the effect of mass insertion in the central honeycombs on the bandgap characteristics of the metastructure. The metastructure is chosen as per the ease of mass insertion and enables us to observe changes in the attenuation band, which are influenced by local resonances occurring at frequencies that vary with the insertion location. Metallic hexagonal inserts are made up of structural steel (as shown in Supplementary Fig. \textcolor{blue}{S5}), and their detailed properties and composition are shown in Supplementary Table \textcolor{blue}{S3}. Given that the lattices are fabricated through 3D printing using Polylactic Acid (PLA), and subsequently metallic inserts of much heavier density are used inside the lattice, these inserts function as local resonators due to the density difference between PLA and steel. These characteristics make the metastructure well-suited for vibration control applications. The presence of inserts significantly influences the wave transmission patterns. Furthermore, the strategic placement of the inserts is considered in such a manner that the inserts themselves make a pattern inside the lattice structure. By changing their location, we can control the bandgap for planar excitation in 1D. In Figure \ref{fig:experimental_validation}(a), we present the effect of random vibration through the simple lattice, while in Fig. \ref{fig:experimental_validation}(b)  we explore the effect of wave transmission in broken symmetry of diamond pattern where the middle row of the lattice is fully filled while the top and bottom row is filled with metallic inserts with a definite pattern. In Fig. \ref{fig:experimental_validation}(c), the effect of wave transmission in the cross-shape pattern is shown where the inserts are arranged alternatively in the middle row of the lattice and the top and bottom rows are completely filled. In Fig. \ref{fig:experimental_validation}(d) lattices are filled in such a manner that it forms an arrow indicating the right direction. Figure \ref{fig:experimental_validation}(e), represents the diamond pattern where the middle row is fully filled; however, the top and bottom rows are alternatively filled. In Fig. \ref{fig:experimental_validation}(f), the top and bottom rows are fully filled and the middle row is fully empty. In Fig. \ref{fig:experimental_validation}(g), the top and bottom rows are fully filled; however, the middle row is filled such that there is a spacing of two units between the consecutive inserts. In Fig. \ref{fig:experimental_validation}(h), the top row is alternatively filled starting with a blank cell while the bottom row is also alternatively filled starting with a filled cell, and the middle row is fully filled except for the first and the last cell representing upward pointing arrow. In Fig. \ref{fig:experimental_validation}(i), the top and bottom rows are alternatively filled while the middle row is filled with two-unit spacing. Through the study of dynamic response corresponding to harmonic excitation for all these combinations, we aim to find out which pattern location is more sensitive and affects more at a specific frequency range of wave transmission. We observed that generally, the middle row exhibits higher sensitivity compared to the top and bottom rows. Additionally, by arranging different combinations of inserts, we achieved up to 48\% bandgap enlargement, multiple bandgaps, and the ability to shift the bandgap across the desired frequency range. Further, the experiments corresponding to each iteration are carried out to validate our simulation results by using a Laser Doppler Vibrometer (LDV). Random vibration up to 10 KHz is performed using LDS shaker with gain set as unity. Detailed experimental setup can be seen in Fig. \ref{fig:experimental_setup}. As observed in Fig. \ref{fig:experimental_validation}, our experimental results show good agreement with the simulation. The printing pattern, and infill density (presented in Supplementary Table \textcolor{blue}{S2}), also affect the attenuation band.

\section{AI-driven Inverse Design of Metastructures} 
\begin{figure*}[!ht]
\centering
\includegraphics[width=16cm]{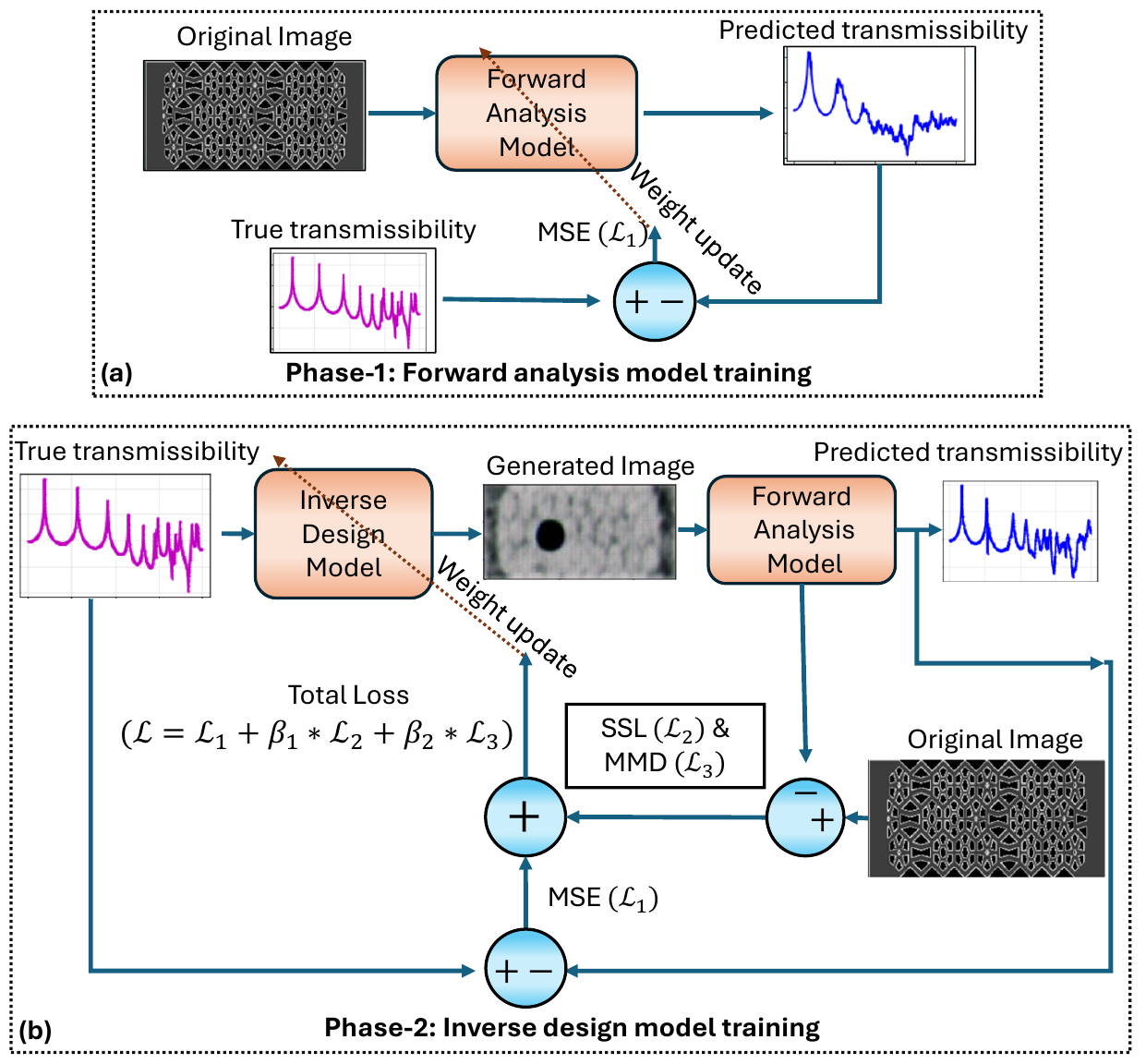}
\caption{AI-based inverse design of metastructures: (a) Training strategy of forward analysis model to predict the transmissibility for a given metastructure (b) Training strategy of inverse design model to predict the metastructure for given transmissibility spectrum.}
\label{fig:AI_Blocks}
\end{figure*}

With the phenomenal expansion potential of the geometry of metastructures, it is possible to use AI efficiently for both forward analysis in predicting the bandgaps as well as inverse design where for a given bandgap a suitable pattern may be obtained.
The complete flow diagram of the proposed AI-based inverse design of metastructures is depicted in Fig. \ref{fig:AI_Blocks}. The entire method comprises two components: (i) the FEM-inspired spatial attention (FSA)-based forward analysis model and (ii) the multiscale Gaussian self-attention (MGSA)-based inverse design model. The forward analysis model acts as a surrogate model for training the inverse design model in the tandem structures of neural networks. The forward analysis model is trained independently using the training samples of metastructure and its transmissibility for the prediction of transmissibility for new metastructures.

\begin{figure*}[ht!]
\centering
\includegraphics[width=\textwidth]{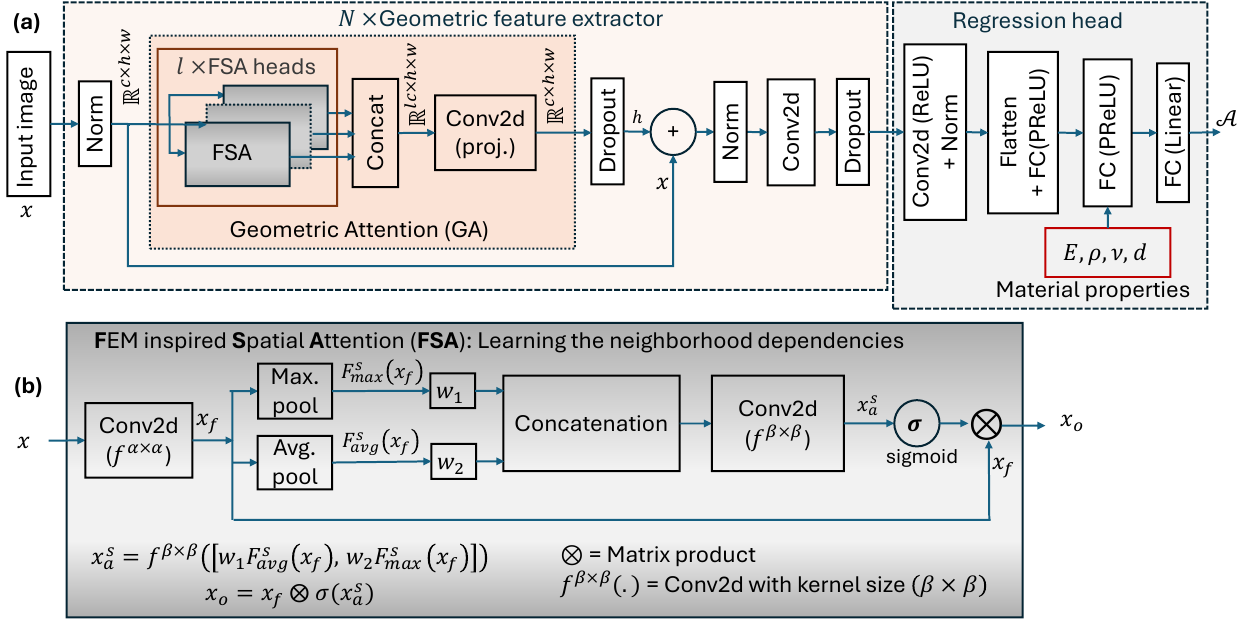}
\caption{\textbf{Detailed block diagram of the FSA-based forward analysis model:} (a) Complete block diagram of the forward model showing the transformation of input metastructure to output transmissibility. The orange box shows the geometric attention (GA) model responsible for capturing the geometrical dependency of vibration mitigation. Grey boxes show the 8 heads of the FEM-inspired spatial attention (FSA) module. (b) Detailed diagram of FSA module. The kernel size of the input convolutional block ($f^{\alpha \times \alpha}$) is different for each FSA head for capturing the inter-dependencies of adjacent connections in the metastructure geometry.}
\label{fig:forwards}
\end{figure*}
\subsection{Forward Analysis Model} The forward analysis model consists of $N$ blocks of geometric feature extractor (GFE), followed by a regression head to process the output of GFE and the material properties to predict the transmissibility of the metastructure. The GFE module is designed on the concept of finite element model (FEM) with the capability of learning the neighborhood dependencies in the complex metastructure geometry having primary, secondary, and tertiary complexity. Each GFE module consists of a multi-scale geometric attention (GA) mechanism having multiple heads of FEM-inspired spatial attention (FSA) blocks. Layer normalization (LN) before the GA and after the summing block is provided to stabilize the training process by reducing internal covariate shifts. Dropout after the GA block is included for the weight regularization. The complete forward model, along with the detailed diagram of FSA is shown in Fig. \ref{fig:forwards}.

Each FSA consists of an input convolutional layer with a different kernel size followed by a spatial feature attention with learnable weights. The kernel sizes of the input convolutional layer in different FSA heads are selected as $\alpha = 1, 3,$ and $5$ depending on the complexity of the input geometry. If $x_f = f^{\alpha\times \alpha}(x)$ is the output of the input convolution, the geometrical feature output based on neighborhood dependencies is obtained using the following equation (\ref{eq:fsa}).
\begin{equation}
    x_a^s = f^{\beta\times\beta}\left(\left[ w_1F^s_{avg}(x_f), \;\; w_2F^s_{max}(x_f)\right]\right) \implies
x_{o} = x_f\otimes\sigma\left(x_a^s\right)\label{eq:fsa}
\end{equation}
where, $f^{\beta\times\beta}(.)$ denotes the convolution with kernel size $\beta$, $\sigma(.)$ denotes the sigmoid operation, and $\otimes$ represents the matrix multiplication. $w_1$ and $w_2$ are learnable weight parameters, initialized with unity. The kernel size $\beta$ is a constant value equal to 7 for our case. Learnable weights in the FSA block are optimized to learn the inter-dependencies of adjacent connections in the spatial dimension. Overall, the whole process attempts to mimic the finite element model to map the vibration spectrum with the metastructure geometry.

The outputs of multiple heads of FSA are concatenated and projected back to have the same number of channels as the input using a 2D convolutional layer. The forward analysis model was implemented with 12 layers of GFE blocks and 8 heads of FSA in each block. The kernel sizes of the input convolution in 8 heads of FSA are selected as $\alpha = 1,\, 1,\, 1,\, 3,\, 3,\, 3,\,5,$ and $5$. This combination of kernel sizes puts more attention on the smallest structures (tertiary) than the secondary and the primary structures. 

The output of the GFE module is processed by a regression head that transforms the geometric features to the transmissibility of dimension $(1000, 1)$. The fully connected (FC) layers in the regression head are associated with Parametric Rectified Linear Units (PReLU) activation. The PReLU activation function facilitates learnable non-linearity to map the highly non-linear nature of the transmissibility. The second FC layer is provided with material properties along with geometric features. These material properties are modulus of elasticity $(E)$, Density $(\rho)$, Poisson's ratio $(\nu)$, and material damping $(d)$. The output of the last-second FC layer is applied to layer normalization (LN) to stabilize the learning. Finally, the last FC layer with linear activation converts the features to the transmissibility having dimension $(1000,\; 1)$.

\begin{figure*}[!ht]
\centering
\includegraphics[width=\textwidth]{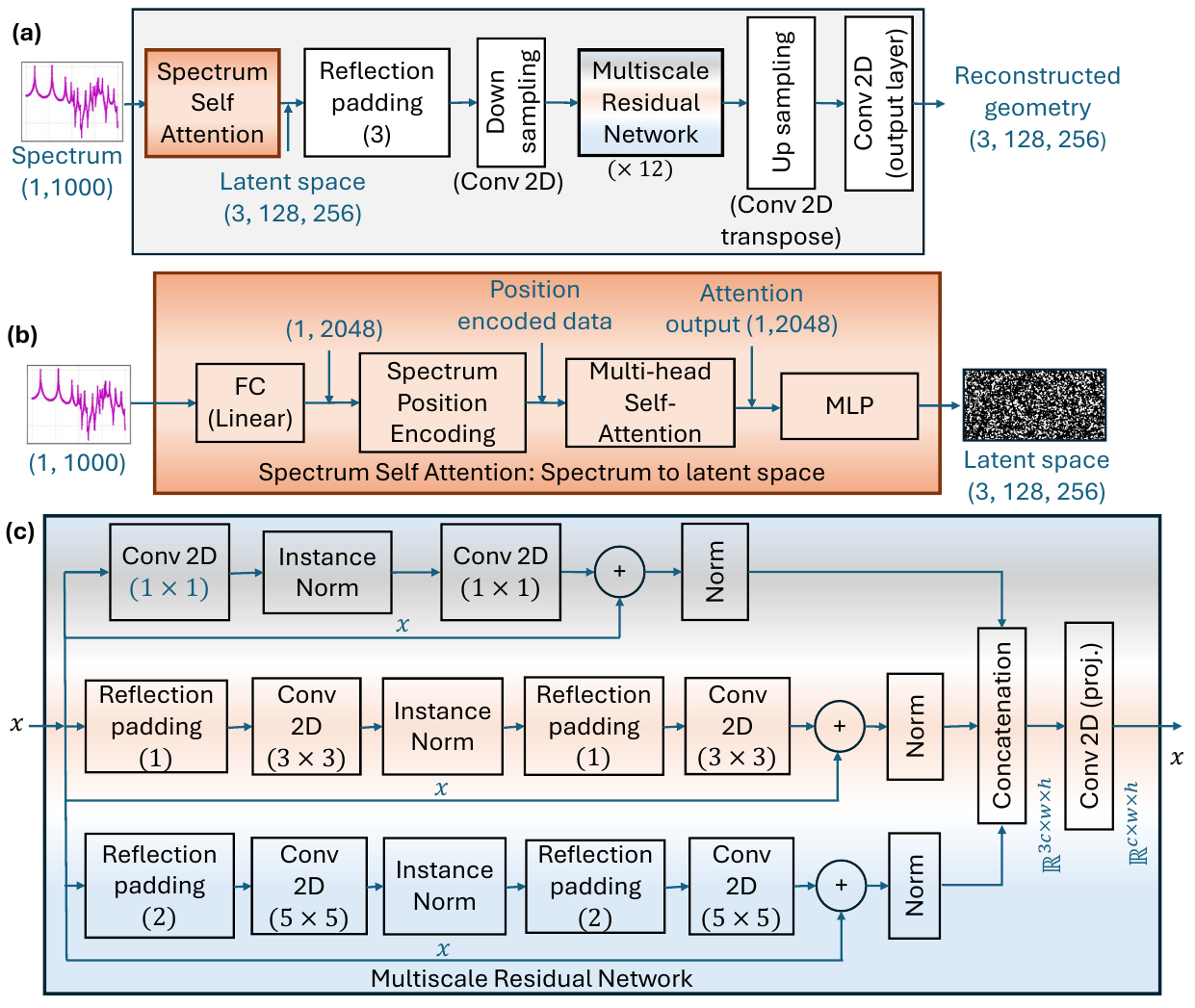}
\caption{\textbf{Detailed block diagram of the inverse design model}. (a) Complete block diagram of inverse design model transforming input transmissibility to metastructure geometry. (b) Detailed block diagram of spectrum self-attention responsible for capturing the sequential dependencies of the transmissibility and generating a rectified latent space. The spectrum position encoding is specially designed using the Gaussian function to emphasize the specific frequency range in the non-linear spectrum. (c) Detailed block diagram of the multiscale residual network. The kernel sizes of the convolutional blocks in three channels are $(1\times1)$, $(3\times3)$, and $(5\times5)$, selected to transform the latent space into the three levels of interlaced structural geometry.}
\label{fig:revModel}
\end{figure*}
\subsection{Inverse Design Model} The inverse design model is developed to obtain the metastructure geometry using transmissibility data for the desired bandgap. The key constituents of the inverse design model are Gaussian self-attention to convert transmissibility to latent space, down sampling network, multiscale residual network to produce interlaced complex structure, and finally upsampling through transpose 2D convolution as shown in Fig \ref{fig:revModel}(a). The Gaussian self-attention is a multi-head self-attention mechanism with a Gaussian function for 1D spectrum position encoding. It captures the sequential dependencies of the transmissibility and generates a rectified latent space of shape  $(3\times128\times256)$, shown in Fig \ref{fig:revModel}(b). Since the transmissibility is the amplitude versus frequency spectrum, a position encoding must be added to represent the absolute position of the amplitude against the log-scale frequency bin. Therefore, a learnable position encoding is defined using sine and cosine of Gaussian functions of log-scale frequency bin to represent the even and odd positions, shown in Equations (\ref{eq:pe}) and (\ref{eq:pe1}), respectively.
\begin{eqnarray}
    \text{PE}(2p) &=& \sin\left(\exp\left({-\frac{(\log(p)-\mu_0)^2}{2\sigma_0^2}}\right)\right)\label{eq:pe}\\
    \text{PE}(2p+1) &=& \cos\left(\exp\left({-\frac{(\log(p)-\mu_0)^2}{2\sigma_0^2}}\right)\right)\label{eq:pe1}
\end{eqnarray}
where $p$ = position index, $\mu_0$ and $\sigma_0$ are the learnable parameters for the dynamic allocation of the mean and standard deviations of the Gaussian distribution for 1D position encoding. $\mu_0$ is initialized with half of the sequence length, whereas $\sigma_0$ is initialized with unity to ensure small spread. The optimization of these parameters enables the Gaussian function to emphasize the specific frequency range in the non-linear spectrum.

The latent space generated by the spectrum self-attention is further extended to a new higher space using reflection padding with a reflection of 3 pixels and then non-linearly down-sampled back to the same dimension using multiple consecutive convolutional layers with ReLU activation.
Now the transformed latent space is passed through a multiscale residual network having three parallel convolutional paths with three different kernel sizes: $(1\times1)$, $(3\times3)$, and $(5\times5)$, shown in Fig. \ref{fig:revModel}(c). These sizes have been selected based on the institution to transform the latent space into the three levels of interlaced structural geometry. The smallest kernel is responsible for learning the tertiary structures, the medium size helps to learn the secondary structural geometry. Similarly, the largest value $(5\times5)$ is responsible for learning the primary geometry. The final output of the multi-scale residual network is the interlaced patterns. It is now up-sampled using three layers of 2D transposed convolution and then transformed non-linearly using 2D convolution with hyperbolic tangent (Tanh) activation to produce the final geometry.

\subsection{Dataset preparation} \label{sec:dataset}
\begin{figure}[ht!]
		\centering
		\includegraphics[width=\textwidth]{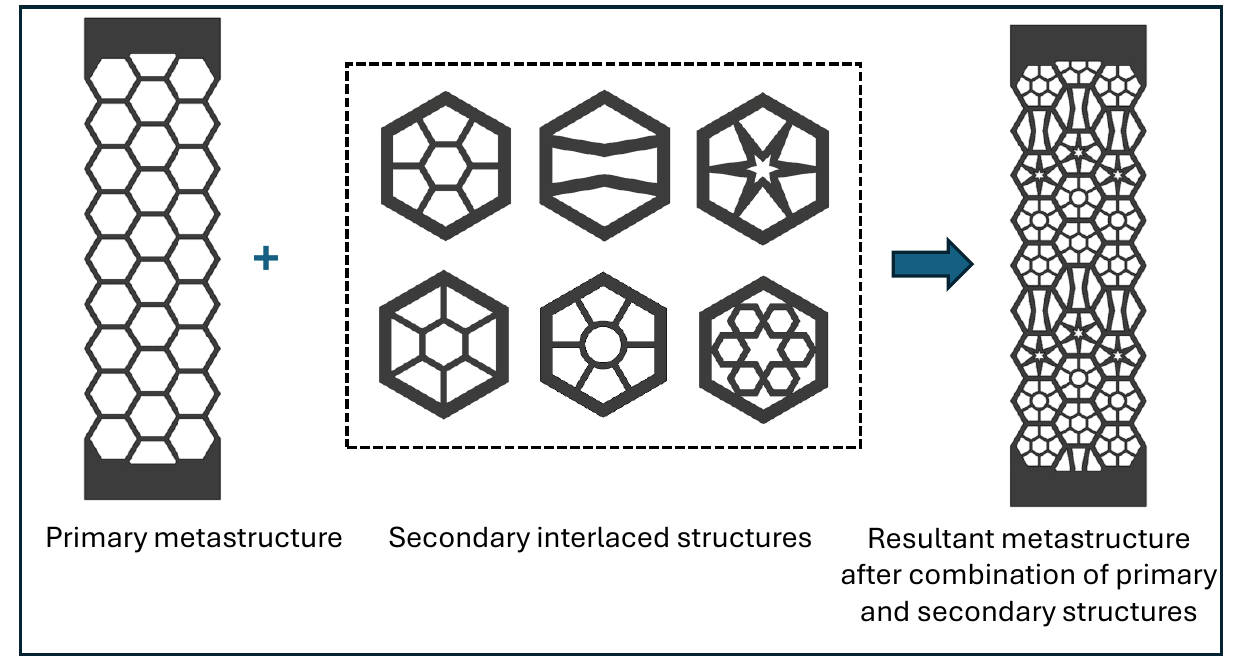}
		\caption{Metastructure of primary geometry as honeycomb and secondary geometry (interlaced structures) including honeycomb, auxetic, star, circle, and hierarchical honeycomb used to design different combinations of metastructures for data set generation using numerical simulation.}
		\label{fig:UnitCell_six}
\end{figure}

A transmissibility dataset was generated for various unique metastructures, having multiscale geometrical designs. These metastructures featured a primary honeycomb structure with different secondary interlaced structures: honeycomb, star, circle, hierarchical honeycomb, and auxetic as shown in Fig.\ref{fig:UnitCell_six}. By integrating these secondary structures with the primary honeycomb, a variety of 720 distinct metastructures having different transmissibilities were created. The transmissibility of each structure was calculated by post-processing the frequency response obtained from Finite Element (FE) simulations conducted in Ansys. Therefore, the training dataset contains 720 samples of metastructure and corresponding transmissibility.

\subsection{Training}
\begin{figure*}[!h]
\centering
\includegraphics[width=\textwidth]{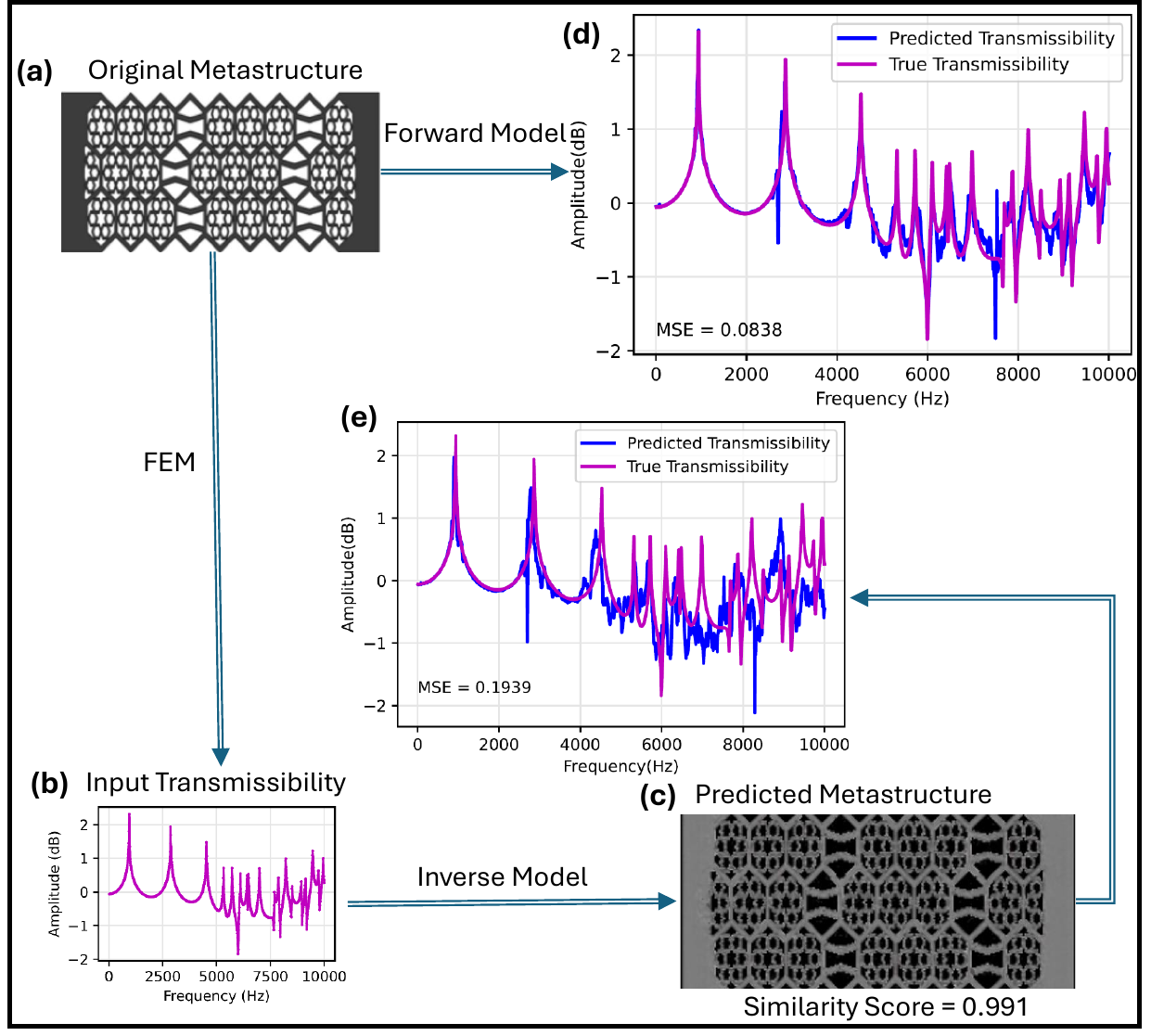}
\caption{\textbf{Validation results for AI-based inverse design of metastructure.} \textbf{(a)} Original metastructure of the test sample. \textbf{(b)} Input transmissibility obtained by FEM for the original metastructure. \textbf{(c)} Metastructure predicted (generated) by the inverse design model using the input transmissibility. The similarity score = 0.991, closer to unity validates the exact match with the original image.  \textbf{(d)} Comparing the transmissibilities of the original metastructure obtained from the forward model and FEM. \textbf{(e)} Comparing the transmissibilities of the generated metastructure obtained from the inverse model and FEM.}
\label{fig:AI_results}
\end{figure*}

The proposed frameworks of forward analysis and the inverse design models were trained on the metastructure-vs-transmissibility dataset as described above. The total samples in the dataset were divided into two portions: training and validation with a ratio of 90:10, resulting in 648 samples in the training set and 72 samples in the validation set. The training of the proposed framework was accomplished in the following two phases. 
\begin{enumerate}[i)]
    \item \textbf{Forward Analysis Model:} In the first phase, the forward analysis model was trained using the training samples to predict the transmissibility of a given metastructures. The training loop has been depicted in Fig. \ref{fig:AI_Blocks}(a). The weight optimization was achieved by minimization of the mean square error between the predicted transmissibility and the true transmissibility. The model was trained for 300 epochs using the Nesterov-accelerated Adaptive Moment Estimation (NAdam) as the optimization algorithm with dynamic allocation of learning rate by the Cosine Annealing learning rate scheduler (details can be found in \cite{nadam} and \cite{cos_sch}).  The comparison of the predicted transmissibility obtained by the forward analysis model and the true transmissibility obtained by the FEM analysis are compared in Fig. \ref{fig:AI_results}.
    \item \textbf{Inverse Design Model:} The trained forward model, obtained from the first phase of the training was used as a surrogate model for the training of the inverse design model as displayed in Fig. \ref{fig:AI_Blocks}(b). The weight optimization was accomplished again using the NAdam optimization algorithm and dynamic allocation of learning rate using the Cosine Annealing learning rate scheduler (details can be found in \cite{nadam} and \cite{cos_sch}).  A composite loss function $(\mathcal{L})$ for the weight update of the inverse design model is defined in Eq. (\ref{lossFn})
    \begin{equation}\label{lossFn}
        \mathcal{L} = \mathcal{L}_1 \;+\; \beta_1*\mathcal{L}_2 \;+\; \beta_2*\mathcal{L}_3
    \end{equation}
    where, $\mathcal{L}_1$ = mean square error (MSE), $\mathcal{L}_2$ = structural similarity loss (SSL), and $\mathcal{L}_3$ = maximum mean discrepancy (MMD). $\beta_1$ and $\beta_2$ are the penalty terms for the inclusion of losses $\mathcal{L}_2$ and $\mathcal{L}_3$. The values of $\beta_1$ and $\beta_2$ were selected as 0.1 and 0.6, respectively. The loss terms MSE, SSL, and MMD are defined as follows:
    \begin{enumerate}[]
        \item \textbf{Mean square error (MSE):} MSE represents the mean deviation of the predicted transmissibility of the generated metastructure and the true transmissibility of the original metastructure. If $a_i \in \mathcal{A}$ and $\hat{a}_i \in \hat{\mathbf{A}}$ are $i^{th}$ data point in the true and predicted transmissibilities,  the MSE is defined in Eq. (\ref{eq:mse}).
            \begin{equation}\label{eq:mse}
                \mathcal{L}_1(\mathcal{A}, \hat{\mathcal{A}}) = \frac{1}{n} \sum_{i=1}^{n} \left ( a_{i}-\hat{a_{i}}\right)^2        
            \end{equation}
        \item \textbf{Structural similarity loss (SSL):} SSL represents the structural misalignment of the generated metastructure and the original metastructure. It is defined in terms of structural similarity measure as shown in Eq. (\ref{eq:ccl})
        \begin{equation}\label{eq:ccl}
            \mathcal{L}_2(\textbf{X},\hat{\textbf{X}}) = 1-\frac{\sigma_{\textbf{X}\hat{\textbf{X}}}}{\sigma_{\textbf{X}}\sigma_{\hat{\textbf{X}}} + c_0}
        \end{equation}
        where $\sigma_{\textbf{X}\hat{\textbf{X}}}$ = covariance between input image $\textbf{X}$ and the generated image $\hat{\textbf{X}}$, $\sigma_\textbf{X}$ = local variance (standard deviation) of the input image $\textbf{X}$, $\sigma_{\hat{\textbf{X}}}$ = local variance (standard deviation) of the generated image $\hat{\textbf{X}}$, and $c_0$ = small constant for division stabilization. The inclusion of SSL with a penalty factor of $\beta_1$ helps the algorithm to force the weight update in the direction to attain the structural similarity between the generated metastructure and the original metastructure. Since the SSL is the opposite of the structural similarity, therefore, $(1-SSL)$ is used to measure the similarity score of the generated image to the original image \cite{ssim}.
        \item \textbf{Maximum mean discrepancy (MMD):} MMD provides the statistical difference (non-parametric distance) between two probability distributions on reproducing kernel Hilbert space (RKHS) \cite{rkhs, 9612606}. Therefore,  MMD can be used for comparing the statistical distributions of the two images by mapping them into a higher-dimensional feature space using a kernel function. If $x_i\in \textbf{X}$ and $\hat{x}_j\in \hat{\textbf{X}}$ are $i^{th}$ and $j^{th}$ data points in the original and the generated image space, respectively, then the MMD of the two distribution in RKHS is defined in Eq. (\ref{eq:mmd})
        \begin{equation} \label{eq:mmd}
            \mathcal{L}_3(\textbf{X},\hat{\textbf{X}}) = \left\Vert {\frac{1}{{n}}\sum \limits _{i = 1}^{{n}} {f \left({{x_i}} \right)} - \frac{1}{{\hat{n}}}\sum \limits _{j = 1}^{{\hat{n}}} {f \left({{\hat{x}_j}} \right)} } \right\Vert _{\mathcal H} 
        \end{equation}
        where, $f(.)$ denotes the kernel function $f:X,X\to {\mathcal H}$, ${\mathcal H}$ be the universal RKHS, $n$ and $\hat{n}$ be the number of pixel points in the original and the generated image $\textbf{X}$ and $\hat{\textbf{X}}$ respectively. The MMD, calculated between the original metastructure and predicted metastructure, with a penalty factor of $\beta_2$, forces the training algorithm to minimize the discrepancies of the generated image compared to the original image.
    \end{enumerate}
The inclusion of SSL and MMD in the loss function helps the training algorithm to optimize the inverse design model to learn the original geometry and makes it less dependent on the performance of the forward analysis model in the tandem network.
\end{enumerate}
\subsection{Validation of AI-model for metastructure design} The metastructure produced by the inverse design model contains colored noises due to training with a small value of batch size (small batch size was chosen due to the limited number of samples in the dataset). Therefore, a noise removal filter is applied to produce a clear image of the obtained metastructure during validation. The algorithm for the noise removal is provided in supplementary material, Section \textcolor{blue}{E}.
The AI-based inverse design of metastructures is validated under two cases:
\subsubsection{Validation on test samples of metastructure:} The comparison of generated metastructure for one test sample from the test dataset (10\% of available 720 samples) is provided in Fig. \ref{fig:AI_results}. Figures \ref{fig:AI_results}(a), (b) show the test metastructure and its transmissibility obtained by the FEM method which is assumed to be true (input) transmissibility for inverse design. The predicted metastructure by the inverse design model exactly matches the original metastructure with the similarity score of 0.991 as shown in Figure \ref{fig:AI_results}(c). Forward analysis is performed on both the original and the predicted metastructures. The predicted transmissibilities are compared with the input (true) transmissibility in Fig. \ref{fig:AI_results}(d) and (e). It can be observed that MSEs for transmissibility obtained from the original metastructure and the predicted metastructure are 0.0834 and 0.1939 respectively. The MSE for predicted transmissibility by the forward analysis on the predicted metastructure is comparatively larger because of the existence of noises in the predicted metastructure. 
\begin{figure*}[!h]
\centering
\includegraphics[width=\textwidth]{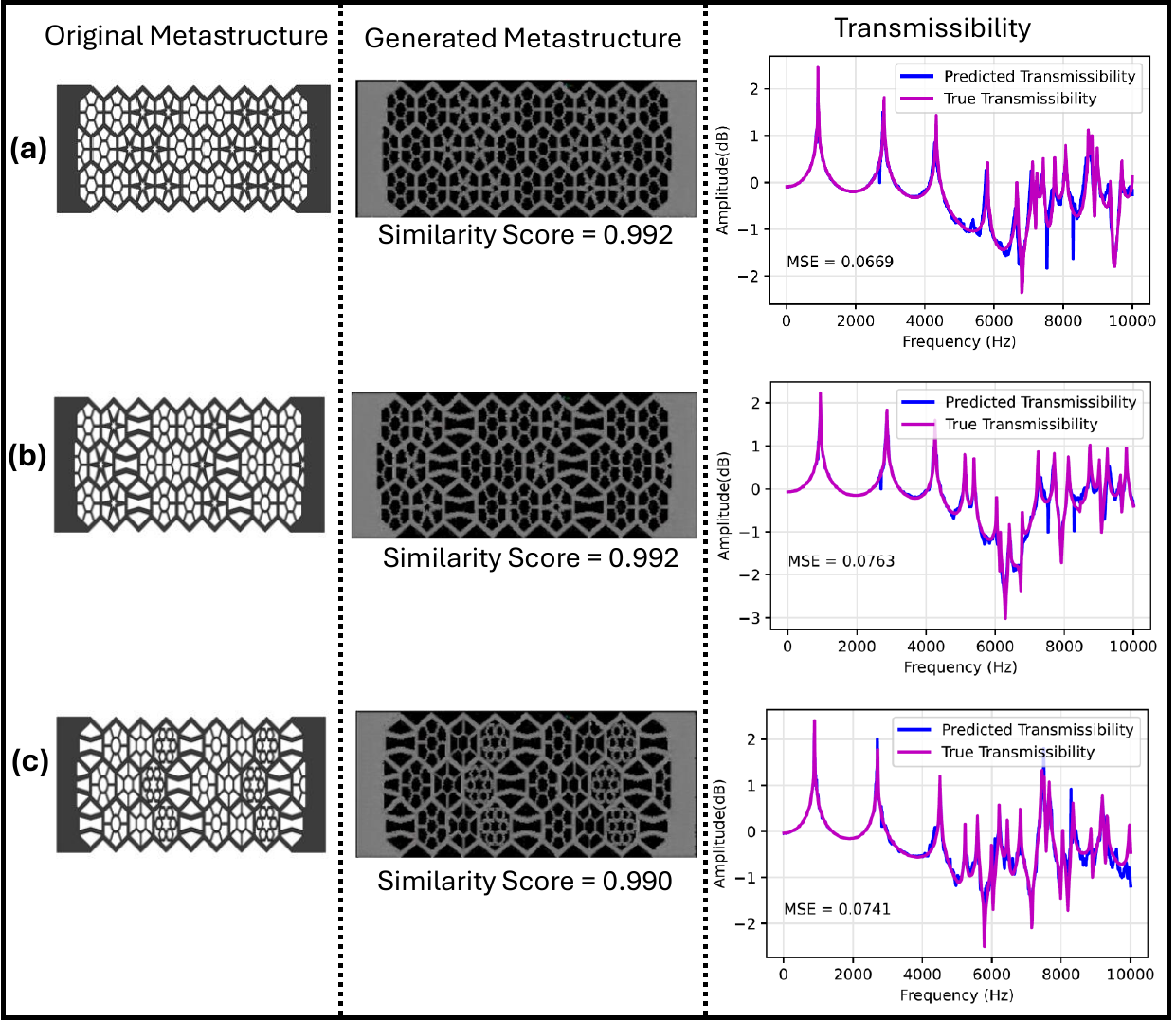}
\caption{Additional validation results on multiscale metastructures with different combinations of the unit cells (Honeycomb as a primary unit cell and auxetic, star, hierarchical honeycomb interlacing as secondary and tertiary).}
\label{fig:AI_results_com}
\end{figure*}

To further demonstrate the performance of the inverse design framework, we have provided comparisons of the generated metastructure with the original metastructure having different combinations of unit cells in Fig. \ref{fig:AI_results_com}. The first column of Fig. \ref{fig:AI_results_com}(a) has honeycomb and star interlacing, Fig. \ref{fig:AI_results_com}(b) has honeycomb, star, and auxetic interlacing, Fig. \ref{fig:AI_results_com}(c) has auxetic, circle, honeycomb, and hierarchical honeycomb interlacing with the similarity score of 0.992, 0.992, and 0.990 respectively. The third column of Fig. \ref{fig:AI_results_com} shows the comparisons of predicted transmissibility of the generated metastructure by the forward analysis model and the true transmissibility evaluated by the FEM on the original metastructure with the MSE of 0.0669, 0.0763, and 0.0741 in respective cases. 
\subsubsection{Validation on tailored transmissibility:} 
\begin{figure*}[!ht]
\centering
\includegraphics[width=\textwidth]{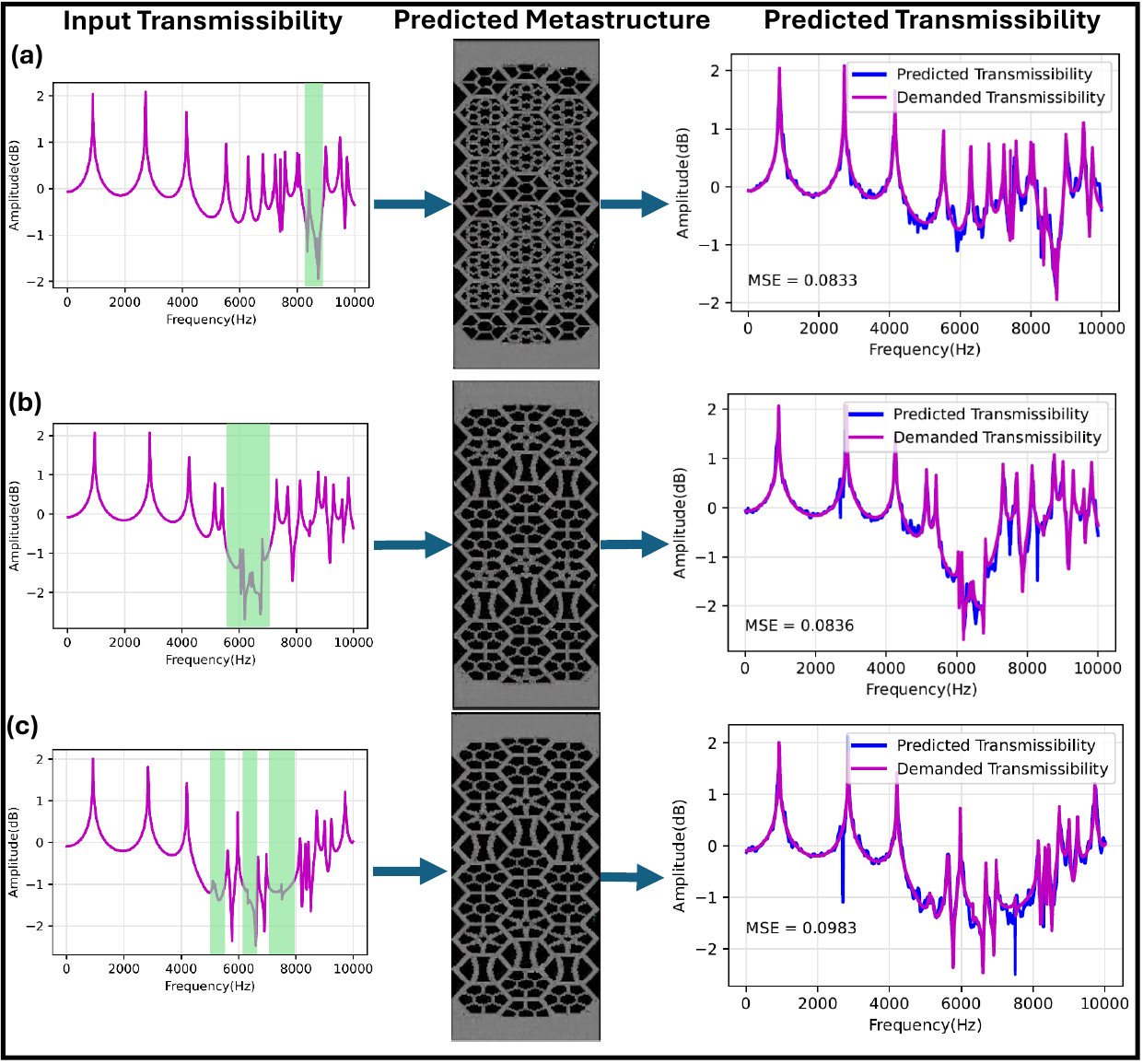}
\caption{\textbf{AI-based inverse design of metastructure for tailored bandgaps.} \textbf{(a)} Transmissibility having tailored bandgap at high frequency. \textbf{(b)} Transmissibility having tailored wider bandgap. \textbf{(c)} Transmissibility having multiple tailored bandgaps ranging from low frequency to high frequency.}
\label{fig:Val_artificial_transs}
\end{figure*}

The AI-based inverse design model is further validated for practical deployment to design a metastructure facilitating the formation of a desired frequency bandgap. Hence, an artificial transmissibility is created with the desired bandgap and fed to the inverse design model. The generated metastructure and the corresponding transmissibilities obtained by the forward analysis model and FEM analysis are shown in Fig. \ref{fig:Val_artificial_transs}. The first column in Fig. \ref{fig:Val_artificial_transs} shows the artificial spectrum created for the tailored bandgap, and the second column presents the metastructures generated from the inverse design model for the input transmissibility with tailored bandgap. The third column compares the predicted transmissibility by the forward model and the true transmissibility evaluated by the FEM on the generated metastructure. 
The predicted metastructure design is the combination of a honeycomb and hierarchical honeycomb substructures as shown in Fig. \ref{fig:Val_artificial_transs}(a). As the attenuation band is in the higher frequency region, hence the inverse design model generates a hierarchical substructure in the generated metamaterial design as previously shown in Fig. \ref{fig:half_power_BW}(a). The MSE between the predicted transmissibility and the input transmissibility is 0.0958. Figures \ref{fig:Val_artificial_transs} (b) and (c) have interlaced structures such as honeycomb, auxetic, circle, and star in predicted metastructure design. The tailored transmissibility has a wider bandgap and multiple bandgaps in figures \ref{fig:Val_artificial_transs} (b) and (c) respectively. Hence the inverse design model generated auxetic substructures in the generated metamaterial design.

\section{Discussion}
 It has been found experimentally and analytically from the previous study that a simple lattice structure has the ability to block waves for a certain frequency range when it passes longitudinally throughout the lattice. Inspired from the architectural patterns, we have designed mechanical metamaterial lattices with diverse interlaced structures. A comprehensive study has been performed through simulation-based analysis using Ansys. Experiments have been conducted to validate the transmissibility property corresponding to different lattice configurations. This work shows the blueprint of metamaterials, including the fact that the effective properties of the lattice are contingent on collective patterns and subpatterns of the unit cells. Furthermore, the tunability of the bandgaps through design resulted in larger and multiple bandgaps occurring in Frequency Response Function (FRF) plots. The reported band gap caused by the lattice structures can be further modified by adding metallic inserts with different combinations of insert patterns. These metallic inserts interact with the geometry and due to the large density difference as compared to lattice material (PLA), they act as local resonators, providing flexibility to adjust bandgaps.  Some significant negative peaks are obtained from the transmissibility plots that signify wave attenuation and show good agreement with the simulation results, as shown in Fig. \ref{fig:experimental_validation}. The discrepancy in some amplitudes and peaks is due to the nonconsideration of structural damping in the simulation. Apart from damping, printing speed, orientation, infill pattern, and infill density of 3D printing are also affected. These findings unveil a deep connection between lattices and band gaps and have broad potential applications in the field of vibration isolation for high-speed trains and in serving as surface waveguides and absorbers.

 The second part of this study on the inverse design of metastructure using geometric attention-based deep architecture is a benchmark in the field of inverse design of complex structure whether it is metastructure for the desired mechanical response or any other complex pattern for desired physical properties. The major challenge in the training of an AI-based inverse design model is the availability of a sufficient number of training samples. The higher the complexity of the structure, the higher the number of samples required for the training. Our proposed model uses FEM to learn the dependencies of adjacent connections in the geometrical structure. Therefore, it can easily extract the information embedded in the complex structure and learn the relationship between the input geometry and the output transmissibility. Further, the multiscale geometrical attention mechanism provides the capability to learn the multiple levels of complexity. With the same concept, multiscale feature transformation using the multiscale residual network enables the inverse design model to reconstruct multiple levels of geometry, like a honeycomb and hierarchical honeycomb substructures.  Therefore, the proposed framework of the geometrical attention-based feature learning and the inverse design can be useful for any other scientific design problem. 

 In summary, our paper reflects the following novelties:
 \begin{itemize}
\item Identification of two unique parameters: interlaced metastructure and fractal order in such patterns followed by their application in 3D printed 2D mechanical metastructure panels.   
\item Further modifications of such metastructures by including local resonators in the form of mass inserts to improve bandgap characteristics through experimental studies.  
\item Creation of novel forward analysis model with a multi-head FEM-inspired spatial attention (FSA) to predict the transmissibility of a metastructure subjected to base excitation. The Multihead FSA efficiently captures the geometrical dependency of vibration mitigation at different scales of structural complexity.
\item The newly developed model is highly efficient as it is trained only on 648 training samples. This results in predicting the transmissibility with the mean square error of 0.08 while in similar literature, ML models were generally trained on 5000 to 40000 data sets to achieve a similar level of accuracy.  
\item Development of a novel MGSA-based inverse design model with 1D spectrum position encoding and a multiscale residual network to produce complex metastructures, given the transmissibility spectrum for a desired frequency bandgap. The proposed inverse design model is found to be much more efficient for the reconstruction of complex metastructure with a similarity score of 0.991. 
\item Further, we have been able to successfully demonstrate the effectiveness of the proposed inverse design model to create complex multiscale metastructures by providing artificial transmissibility for a desired frequency bandgap. We believe this will facilitate the generation of many customizable complex patterns for mechanical metastructures efficiently.
 \end{itemize}



\section{Conclusions}

This research study highlights the translational application potential of the vast range of patterns and motifs developed in the field of architecture into metamaterial-based wave attenuation. Thus, the utilitarian values of aesthetic constructs in the field of vibration and sound engineering may open up new systems for wave control, including bandwidth control, sub-wavelength acoustic imaging, and generation of acoustic black holes. 
Additionally, a systematic investigation process has been created on the basis of root lattice identification, multi-scale lattice formation through a hierarchic system, development of hybrid lattice structure, and the effect of interlacing in the vibration mitigation potential. In general, the simple lattice structures typically act as attenuators, however, this study demonstrates that the interlaced structure offers an enhanced vibration attenuation property depending on the interlace geometry with improved control over the attenuation frequency range. Such effects can be further enhanced in the presence of and through the strategic positioning of the metallic inserts inside the lattice structure, which may result in a significant expansion of the bandwidth. This method has been enhanced with the help of AI-based inverse design models to predict metastructures that exhibit the desired bandgap and the corresponding transmissibility. Although this study concentrates on predominantly honeycomb-shaped interlaced structures, the principle of incorporating mass resonators can be applied to other geometries to manipulate bandgap frequencies. Moreover, the concept may be extended to three-dimensional interlaced structures with diverse lattice and interlace configurations across multiple length scales. Another possibility of future expansion is related to the insertion of smart materials in the lattice structure such that the stiffness and the damping can be controlled over time; thus, transforming it into 4-D metamaterial.

 \section*{Data and code availability}
 All relevant data and codes will be available upon reasonable request from the contact author. The research data are presented either in the main file or in a supplementary file.

\bibliographystyle{elsarticle-num} 
\bibliography{refs}


\appendix

\section*{Acknowledgements} 
	The authors want to thank Mr. Gyanendra Prakesh Tripathi, who provided insight to conduct the experiments, and Mr. Abhishek Kumar Singh, who helped to 3D print the samples. The photo credits for Figure \ref{fig:tomb}(a-f) are given to Chitralekha Bhattacharya and Bipasana Bhattacharya.

\subsection*{Author contribution}
 \textbf{Tanuj Gupta:} Conceived and expanded the idea jointly, modeled all the structures, and designed and performed numerical simulations. Fabricated the samples by 3D printing and carried out related experiments and material testing, documented the results, reviewed the manuscript, critically examined the results, and checked the data, wrote the manuscript. \textbf{Arun Kumar Sharma:} Concived the idea of AI-based inverse design of metastructures, and implemented a new strategy of feature learning using multiscale GA and the idea of reproducing the geometry with multiple levels of complexity using the multiscale residual network. Code Implementation, training, and validation of the AI models and writing the AI-based inverse design in the manuscript. Critically examined the manuscript. \textbf{Ankur Dwivedi:} Conceived and expanded the idea jointly for the inverse design of metastructures, critically examined the results, checked the data, wrote and reviewed the manuscript. \textbf{Vivek Gupta:} Expanded the idea jointly, reviewed the manuscript, critically examined the results, checked the data, and wrote the manuscript. \textbf{Subhadeep Sahana:}  Carried out numerical simulations to generate synthetic data for training the AI model and critically analyzed the paper for logical conclusions. \textbf{Suryansh Pathak:} Conducted the experiments, performed numerical simulations, and checked the data. \textbf{Ashish Awasthi:}  Carried out numerical simulations to generate synthetic data for training the AI model. \textbf{Bishakh Bhattacharya:} Conceived the primary idea, reviewed the manuscript, critically examined the results, checked the data, wrote the manuscript, and acquired funding. 

 \subsection*{Competing interest}
The authors declare no competing interests.



\end{document}